\documentclass[10pt,twocolumn,letterpaper]{article}

\usepackage{wacv}
\usepackage{times}
\usepackage{epsfig}
\usepackage{graphicx}
\usepackage{amsmath}
\usepackage{amssymb}

\usepackage{multirow}
\usepackage{adjustbox}
\usepackage[labelsep=period,font=small]{caption}

\usepackage[pagebackref=true,breaklinks=true,letterpaper=true,colorlinks,bookmarks=false]{hyperref}

\wacvfinalcopy % *** Uncomment this line for the final submission

\def\wacvPaperID{116} % *** Enter the wacv Paper ID here

\ifwacvfinal\fi
\begin{document}

\newcommand*\samethanks[1][\value{footnote}]{\footnotemark[#1]}

%%%%%%%%% TITLE
\title{A Benchmark on Tricks for Large-scale Image Retrieval}

% Authors at different institutions
\author{Byungsoo Ko \thanks{These authors contributed equally to the work.} \\
NAVER/LINE Vision\\
{\tt\small kobiso62@gmail.com}
\and
Minchul Shin \samethanks \\
Search Solutions Inc.\\
{\tt\small min.stellastra@gmail.com}
\and
Geonmo Gu \hspace{2cm} Heejae Jun \hspace{2cm} Taekwan Lee \hspace{2cm} Youngjoon Kim \\
NAVER/LINE Vision\\
{\tt\small \{geonmo.gu, heejae.jun, taekwan.lee, kim.youngjun\}@navercorp.com}
}

\maketitle
\ifwacvfinal\thispagestyle{empty}\fi

%%%%%%%%% ABSTRACT
\begin{abstract}
Many studies have been performed on metric learning, which has become a key ingredient in top-performing methods of instance-level image retrieval.
Meanwhile, less attention has been paid to pre-processing and post-processing tricks that can significantly boost performance.
Furthermore, we found that most previous studies used small scale datasets to simplify processing.
Because the behavior of a feature representation in a deep learning model depends on both domain and data, it is important to understand how model behave in large-scale environments when a proper combination of retrieval tricks is used.
% generally performed the evaluation on small-scale datasets due to its easiness to process.
% Since the behavior of a feature representation of a deep learning model has a dependency on domain and data, it is essential to get a sense of how the model would behave in large-scale environment followed by a proper combination of retrieval tricks.
In this paper, we extensively analyze the effect of well-known pre-processing, post-processing tricks, and their combination for large-scale image retrieval.
We found that proper use of these tricks can significantly improve model performance without necessitating complex architecture or introducing loss, as confirmed by achieving a competitive result on the \textit{Google Landmark Retrieval Challenge 2019}.
\end{abstract}

\begin{figure}[t]
\begin{center}
\begin{adjustbox}{width=1.1\columnwidth,center}
\includegraphics[width=1.0\linewidth]{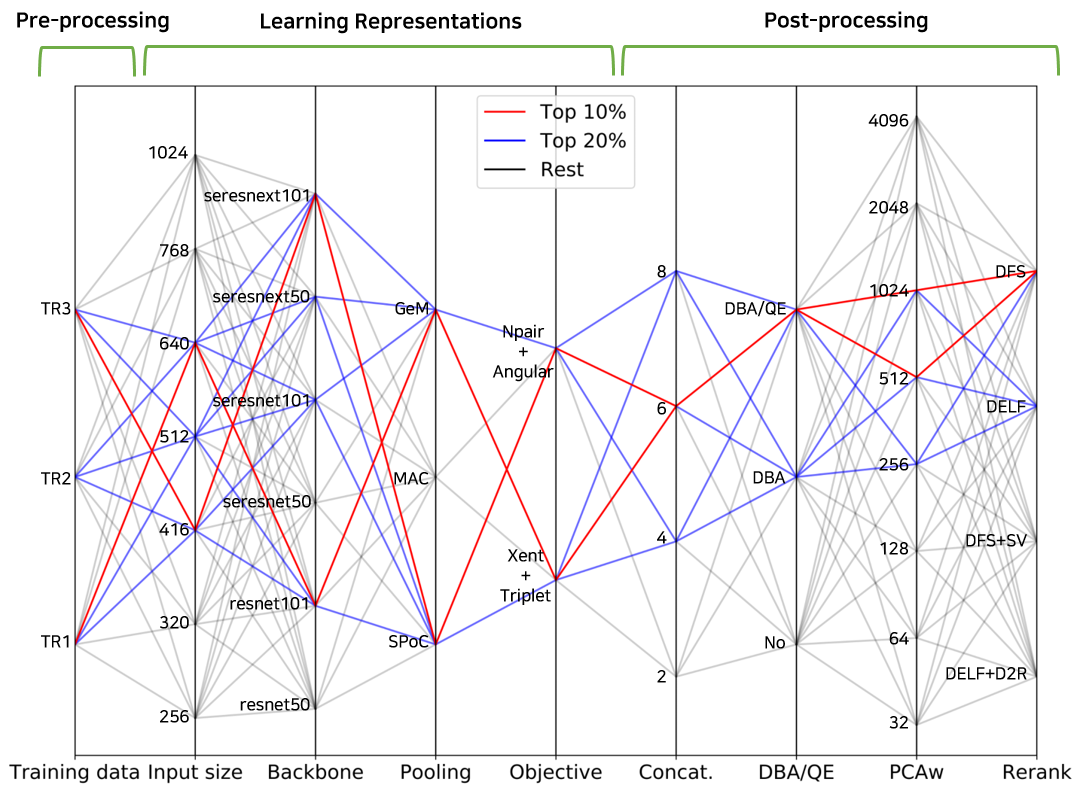}
\end{adjustbox}
\end{center}
   \caption{Possible combinations of pipeline for large-scale image retrieval task. The pipeline consists of pre-processing, learning representations, and post-processing. The routes shown in red and blue indicate top 10\% and top 20\% of combinations, respectively, in terms of performance.}
\label{fig:teaser}
\end{figure}

%%% Introduction
\section{Introduction}
With recent advances in metric learning techniques, there have been active studies~\cite{noh2017large,gordo2017end,jun2019combination,wang2017deep,sohn2016improved} that aim to improve the model performance on image retrieval.
Most previous methods have focused on learning a good representation by carefully designing loss functions~\cite{2019arXiv190705007S,yu2018hard} and architecture~\cite{li2018harmonious}, and evaluated their performance against relatively small-scale datasets such as \textit{Oxford5k}~\cite{philbin2007object}, and \textit{Paris6k}~\cite{philbin2008lost}.
These datasets are well-structured, reliable, and sufficiently small to be processed easily.
However, high performance on a small dataset does not guarantee that the discovered model is generalized.
Although a few studies~\cite{magliani2018efficient} have attempted to investigate large-scale image retrieval using datasets such as \textit{Oxford105k}~\cite{philbin2007object} and \textit{Paris106k}~\cite{philbin2008lost}, we found that most of the public large-scale datasets are not truly large; instead they typically consist of a small number of query and index images, and a large number of irrelevant images used as distractors.
Moreover, we have seen few works that evaluate performance against a dataset that includes more than 100K queries and is actually large-scale in terms of both query and index size.
The reason is simple: querying more than 100K images from among a large number of indexed images is extremely time-consuming and computationally expensive.

On the other hand, the tricks~\cite{turcot2009better,arandjelovic2012three,iscen2017efficient,noh2017large} used during pre-processing and post-processing also have a significant impact on retrieval performance.
F. Radenovic \textit{et al.}~\cite{radenovic2018revisiting} reported enormous improvement in performance by applying several retrieval tricks before and after the main processing.
Despite the importance of such tricks, we found few papers that discuss the best-performing pre-processing and post-processing tricks, particularly for large-scale environment.

In this paper, we aim to analyze how retrieval performance varies on large-scale datasets when different types of pre-processing and post-processing tricks are used in combination.
To do so, we design a very simple model that uses no difficult engineering (such as network surgery) and analyze the effect of the well-known retrieval tricks on the performance in a stepwise fashion.
Extensive experiments were conducted for all possible combinations of tricks and models with various hyper-parameters, as shown in Figure~\ref{fig:teaser}.
We found that proper usage of well-known retrieval tricks can significantly improve the overall performance even when using a simple inference model.

\vspace{-4mm}
\paragraph{Main contributions.} Our contribution is two fold:
(1) We analyze the effect of pre-processing, post-processing tricks, and their combination on large-scale image retrieval with extensive experiments.
(2) We show that competitive improvement can be achieved by properly combining well-known retrieval tricks. Our pipeline was ranked 8-th place in the Google landmark retrieval challenge 2019~\cite{glc2}.
% (1) We study the behavior of a feature representation of a deep learning model for retrieval on the large-scale environment.
% (2) We show that a competitive improvement can be achieved by properly combining well-known retrieval tricks including pre-processing and post-processing. Our pipeline ranked at \textbf{BLINDREVIEW}-th place in the Google landmark retrieval challenge 2019~\cite{glc2}.

% (1) We study the effect of well-known retrieval tricks when they are mutually combined for large-scale image retrieval task.
% (2) With only simple and well-known techniques, our pipeline achieved \textbf{BLINDREVIEW}-th place in the google landmark retrieval challenge 2019~\cite{glc2}.

%%% Models and Pipeline
\section{Pre-processing Tricks}
We used two large-scale datasets for the experiments: \textit{Google landmark dataset (GLD) v1}~\cite{noh2017large} and \textit{GLD v2}~\cite{gldv2} from the 2018 and 2019 Google landmark challenges, respectively.
Detailed statistics for the datasets are shown in Table~\ref{Table:dataset}.
When dealing with large-scale datasets, several unexpected problems can arise.
The first issue is how to eliminate noise from such large-scale datasets.
As a dataset becomes larger, more noise which functions as distractors could be included.
Therefore, noise should be carefully eliminated because the quality of training data significantly affects the model performance~\cite{gordo2016deep,gordo2017end}.
The second issue is how to quickly evaluate model performance through enough trials to perform a large number of experiments.
Different from other famous landmark datasets~\cite{radenovic2018revisiting}, the GLD datasets contain both more than 0.7 million index images and more than 0.1 million query images.
This amount of data requires a significant amount of time and memory to conduct even a single evaluation.
In this section, we describe how to circumvent these issues by removing noise images and constructing a small-scale validation set.

%% Table: Dataset statistics
\begin{table}[t!]
\begin{adjustbox}{width=1.0\columnwidth,center}
\begin{tabular}{c c c c c c c c}
\hline
\multicolumn{2}{c}{\multirow{2}{*}{Dataset}} & \multicolumn{2}{c}{Raw}  & \multicolumn{4}{c}{Pre-processed} \\ \cline{3-8} 
\multicolumn{2}{c}{}                         & GLD v1      & GLD v2      & TR1   & TR2  & TR3  & Valid.  \\
\hline \hline
\multirow{2}{*}{Train}        & Class        & 14.9K   & 203K  &   22.8K    &   22.9K   &   58.3K   &   -   \\
                              & Image         & 1.19M & 4.13M &   1.67M   &   1.68M   &   2.0M   &   -   \\ \hline
\multirow{2}{*}{Test}         & Class        & -     & -     &   -       &   -      &   -      &   2.4K   \\
                              & Image         & 115K  & 117K  &   -       &   -      &   -      &   20.4K   \\ \hline
\multirow{2}{*}{Index}        & Class        & -     & -     &   -       &   -      &   -      &   2.5K   \\
                              & Image         & 1.09M & 0.76M &   -       &   -      &   -      &   67.4K   \\ \hline
\end{tabular}
\end{adjustbox}
\caption{Number of images and classes for each version of GLD, and pre-processed training sets (TR1, TR2, and TR3) and validation set (Valid.). Note that the images that could not be downloaded were excluded.}
\label{Table:dataset}
\vspace{-2mm}
\end{table}

\subsection{Dataset Cleaning for Training}
By cleaning the noise from the training set, we aim to maximize inter-class variation and minimize intra-class variation.
To resolve this without supervision, we used a clustering technique, the density-based spatial clustering of applications with noise (DBSCAN)~\cite{ester1996density}, which can be replaced by any clustering algorithm~\cite{ray1999determination,reynolds2015gaussian}.
Based on the generated clusters, three different types of clean datasets (\textit{TR1, TR2, and TR3}) were constructed.
The detailed procedures for constructing each dataset are described below.

\vspace{-4mm}
\paragraph{TR1.}
We noticed by visual inspection that the training set of GLD v1 is clean and reliable, so we used the dataset as is for training.
To obtain a semi-supervised learning effect~\cite{babakhin2019semi, li2018naive}, we added virtual classes to the training set.
These virtual classes are the clusters from the test and index sets of GLD v1.
First, we trained a baseline model using GLD v1 and extracted features of the test and index set of GLD v1.
Then, DBSCAN was applied to generate clusters, where each cluster was assigned as a new virtual classes.
We call the result TR1 for clarity.

\vspace{-4mm}
\paragraph{TR2.}
In the index set of GLD v2, there are many distractor images that are not a landmark, such as documents, portraits, and nature scenes.
We aimed to use these noises for the training phase so that the model could increase the distance between real landmarks and noises in an embedding space.
For this, we performed the same procedure of clustering but also picked several distractor clusters as virtual classes.
These distractor classes were combined with TR1, and we call the combined dataset TR2.

\vspace{-4mm}
\paragraph{TR3.}
The training set of GLD v2 has more classes and images than GLD v1 does.
At the same time, it contains a large number of noise classes and images.
To address this, we first removed the nature scenes using a simple binary classifier trained on the \textit{Open Images Dataset}~\cite{krasin2017openimages} and the \textit{iNaturalists Dataset}~\cite{van2018inaturalist}.
Then, the images in each class were clustered so that the noise could be excluded as outliers.
When multiple clusters were found in a class, we chose the largest cluster and discarded the others.
Moreover, the classes from the training set of GLD v2 that were duplicated in TR2 were also excluded by querying the images of each training class.

% Figure: before / after DBSCAN
\begin{figure}[t!]
\begin{center}
\includegraphics[width=1.0\linewidth]{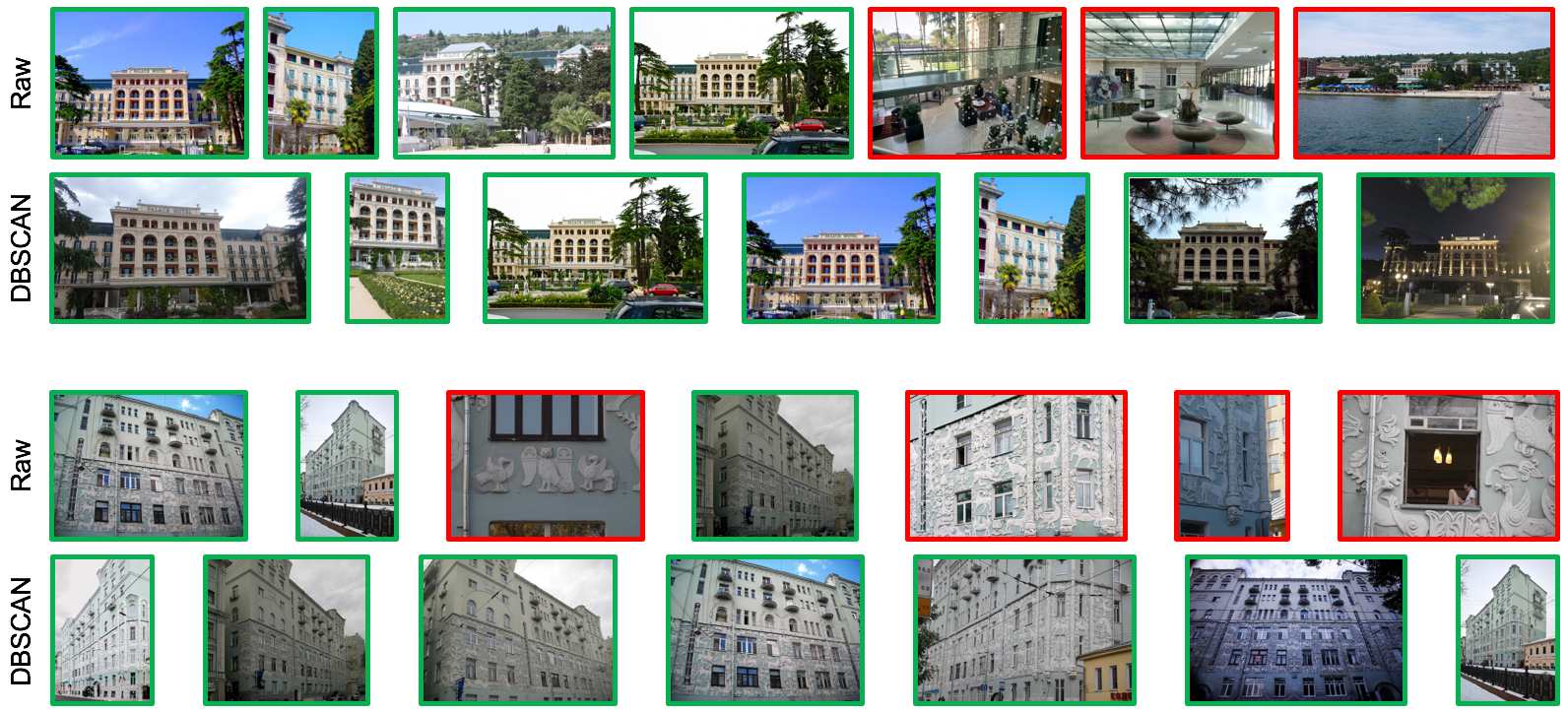}
\end{center}
\caption{Before and after dataset cleaning using DBSCAN. The first row is raw images, and the second row is clean images. The green-bordered images are gound-truth images, and the red-bordered images are noise. The image that is partial or taken from inside is considered as distractors.}
\label{fig:dbascan}
\end{figure}

%% Table 4: Dataset cleaning before & after
\begin{table}[t!]
% \footnotesize
\centering
\begin{tabular}{c c c c c}
\hline
Backbone    &   Train Set   &   Valid.   &   Public   &   Private   \\
\hline \hline
SE-ResNet50   &  GLD v1 &   84.48   &   15.75   &   18.02    \\
SE-ResNet50   &   TR1   &   83.90   &   16.63   &   18.52    \\
SE-ResNet50   &   TR2   &   83.96   &   16.29   &   18.64    \\
SE-ResNet50   &   TR3   &   \textbf{85.15}   &   \textbf{17.94}   &   \textbf{19.90}    \\
\hline
\end{tabular}
\caption{Performance (mAP@100) evaluation of validation set and submission with different training set.}
\label{Table:cleaning}
\vspace{-2mm}
\end{table}

\subsection{Small-scale Validation}
Saving time for each evaluation run is important because it determines how many times we can run trials to validate our hypotheses.
Furthermore, the validation set should reflect the characteristics of the test data as much as possible to prevent misguided interpretation.
To obtain substantially smaller dataset, we sampled about 2\% of the training images from GLD v2 and divided the sample into test and index sets.
We included a virtual class from a noise cluster because the test set of GLD v2 includes a number of distractor images.
In this way, we expect that the distribution of the validation set will be similar to the test and index set of GLD v2.
We report the validation score along with the submission score because the best performing hyper-parameters of each model were explored on the basis of the validation score.

\subsection{Experimental Results}
The Figure~\ref{fig:dbascan} shows an example of how the noisy ground-truth labels have become clean after the clustering.
The raw dataset includes images taken from inside, outside, and even partial viewpoints from within the same landmark.
These kinds of datasets with large intra-class variation may interfere with learning proper representations in the model, especially when a pair-wise ranking loss is used.
Moreover, images of nature scenes also make the training process hard as they have a little iter-class variation.
After refining the raw dataset by choosing a big cluster and removing distractors, we obtained a cleaned dataset.

We performed experiments with a N-pair + Angular model by differentiating the training set and validated each performance.
The input size was 256 $\times$ 256 px for the training phase, 416 $\times$ 416 px for the inference phase, with 1024-dimensional embedding.
As shown in Table~\ref{Table:cleaning}, training with TR1, which contains virtual classes from the test and index sets, improves the model performance by using unlabeled data when the original training set is not helpful anymore.
The model trained with TR2 gives performance similar to the model trained with TR1 because the number of data and classes is not noticeably different.
Because the training set of GLD v2 is quite noisy, we could not train a model from the raw dataset.
Using TR3, which includes a cleaned training set from GLD v2, further improved the performance.
Overall, both validation and submission performance were improved by clustering the data more finely and including more images.
The experiments also showed that the validation set is suitable for use as the performance on validation and submission have similar patterns.

\section{Learning Representations}
In many cases, state-of-the-art performance is obtained by using a novel design of architecture or loss function~\cite{dai2018batch,kim2018attention,li2018harmonious}.
Although such methods may show comparable results on GLD, we intentionally designed a very simple model because we are more interested in the effect of pre-processing and post-processing tricks on large-scale datasets than in the inference model itself.
Nevertheless, it is still interesting to figure out which combination of commonly used pooling methods and objectives work best for the task.
Therefore, we trained multiple models with different types of pooling methods and loss functions expected to be helpful for the feature ensemble in post-processing.

\subsection{Pooling}
Despite the simplicity of implementation, the pooling method on the feature map from the last layer affects the model performance fairly strongly~\cite{babenko2015aggregating,tolias2015particular,radenovic2018fine,tolias2016rmac}.
For this reason, in~\cite{jun2019combination} extensive experiments were performed to find the optimal combination of pooling methods.
However, here, we found that the best pooling method depends on the domain.
Ultimately, three different methods (SPoC~\cite{babenko2015aggregating}, MAC~\cite{tolias2015particular}, and GeM~\cite{radenovic2018fine}) were used for training in this paper.

%% Table 1: Single Model Performance
\begin{table*}[t!h!]
%\footnotesize
\centering
\begin{tabular}{cccccccccc}
\hline
Index & Train Set & Objective     & Input & Backbone      & Pooling & Dim. & Valid. & Public & Private \\ \hline\hline
0     & TR1       & Xent+Triplet  & 640   & ResNet101     & GeM     & 1024 & \textbf{85.78}  & \textbf{19.27}  & 21.79   \\
1     & TR1       & Xent+Triplet  & 640   & ResNet101     & SPoC    & 1024 & 84.25  & 18.19  & 19.91   \\
2     & TR1       & Xent+Triplet  & 640   & SE-ResNeXt50  & GeM     & 1024 & 85.35  & 19.04  & 21.51   \\
3     & TR3       & Xent+Triplet  & 640   & SE-ResNeXt50  & GeM     & 1024 & 85.63  & 19.20  & 21.20   \\
4     & TR3       & Xent+Triplet  & 640   & ResNet101     & GeM     & 1024 & 85.68  & 19.12  & 21.37   \\
5     & TR4       & N-pair+Angular & 416   & SE-ResNet50   & SPoC    & 1024 & 85.52  & 16.83  & 19.76   \\
6     & TR4       & N-pair+Angular & 416   & SE-ResNeXt50  & SPoC    & 1024 & 85.68  & 17.90  & 20.68   \\
7     & TR4       & N-pair+Angular & 416   & SE-ResNet101  & SPoC    & 1024 & 85.60  & 19.08  & 20.65   \\
8     & TR4       & N-pair+Angular & 416   & SE-ResNeXt101 & SPoC    & 1024 & 85.68  & 19.01  & \textbf{22.14}   \\ \hline
\end{tabular}
\caption{Performance (mAP@100) evaluation for top 9 single models.}
\label{Table:single_model_performance}
\end{table*}

\begin{figure}[t]
\begin{center}
\includegraphics[width=1.0\linewidth]{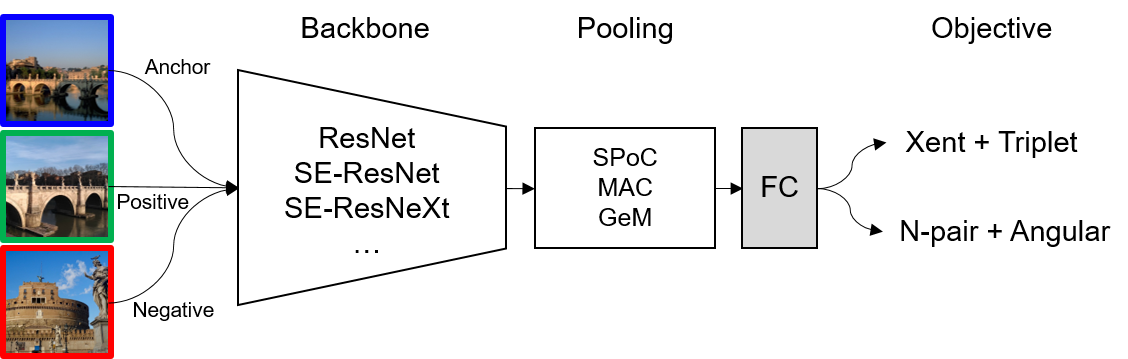}
\end{center}
   \caption{The architecture of a simple network for learning representations.}
\label{fig:training_model}
\vspace{-2mm}
\end{figure}

\subsection{Objectives}
We designed two types of objectives by combining two well-known loss functions.
We expect that differentiating the objectives will enrich the variation of the model representation, which is helpful for the feature ensemble.

\vspace{-4mm}
\paragraph{Xent + Triplet.}
Triplet loss is one of the simplest ranking loss, but we found it offered comparable performance to other loss functions when combined with a classification loss such as cross-entropy (Xent) loss~\cite{jun2019combination,berman2019multigrain}.
In this case, the number of instances becomes the number of classes to classify, and the triplet pair is sampled from the given minibatch using hard example mining.

\vspace{-4mm}
\paragraph{N-pair + Angular.}
N-pair + Angular loss function is a combined form of pair-wise ranking losses: N-pair~\cite{sohn2016improved} and Angular~\cite{wang2017deep}.
Angular loss can be easily integrated into N-pair loss, and we expect that combining the distance-based loss with the angle-based loss will renders the objective function more robust against large variations in feature maps~\cite{wang2017deep}.

\subsection{Training a Single Model}
On the basis of the aforementioned pooling methods and objective functions, we design a very simple model by replacing the components in Figure~\ref{fig:training_model}.
We used ResNet~\cite{he2016deep} applied tricks from Xie \textit{et al.}~\cite{xie2018bag}, SE-ResNet~\cite{he2016deep, hu2018squeeze}, and SE-ResNext~\cite{xie2017aggregated, hu2018squeeze} as a backbone to enhance the structural variation.
A fully-connected layer and l2-normalization are used after pooling the output feature maps.
The number of possible combinations to produce a single model comes to $backbone(6) \times pooling(3) \times objective(2)=36$ in total.

\subsection{Experimental Results}
We trained multiple models with various combinations of training data, input size, backbone, pooling method, and objective.
For the N-pair + Angular model, we used a 256 $\times$ 256 px input size for the training phase and a 416 $\times$ 416 px input size for the inference phase.
For the Xent + Triplet model, the input size was 320 $\times$ 320 px for the training phase and 640 $\times$ 640 px for the inference phase.
The output dimensionality for both models was 1024.
The performance of the top 9 models are reported in Table~\ref{Table:single_model_performance}.

\vspace{-4mm}
\paragraph{Pooling.}
The performance of each pooling method can differ with the characteristics of dataset and model~\cite{boureau2010theoretical}.
Within the three pooling methods, MAC showed the worst performance among all model combination.
The best performing pooling methods differed with the objective.
SPoC showed the best performance for N-pair + Angular models, while GeM was the best pooling method for Xent + Triplet models.

\vspace{-4mm}
\paragraph{Objectives.}
As shown in Table~\ref{Table:single_model_performance}, models with two different objectives had similar performances but different tendency when it comes to the training data.
Unlike the N-pair + Angular models, the Xent + Triplet models showed the best performance with TR1.
With TR3, the Xent + Triplet model could not be trained properly owing to fluctuation of the Xent loss.
This can be because the Xent loss is sensitive to the quality of the dataset, as it may obtain a small number of duplicated classes during the refinement process.

\begin{figure*}[t!h!]
\begin{center}
\includegraphics[width=1.0\textwidth]{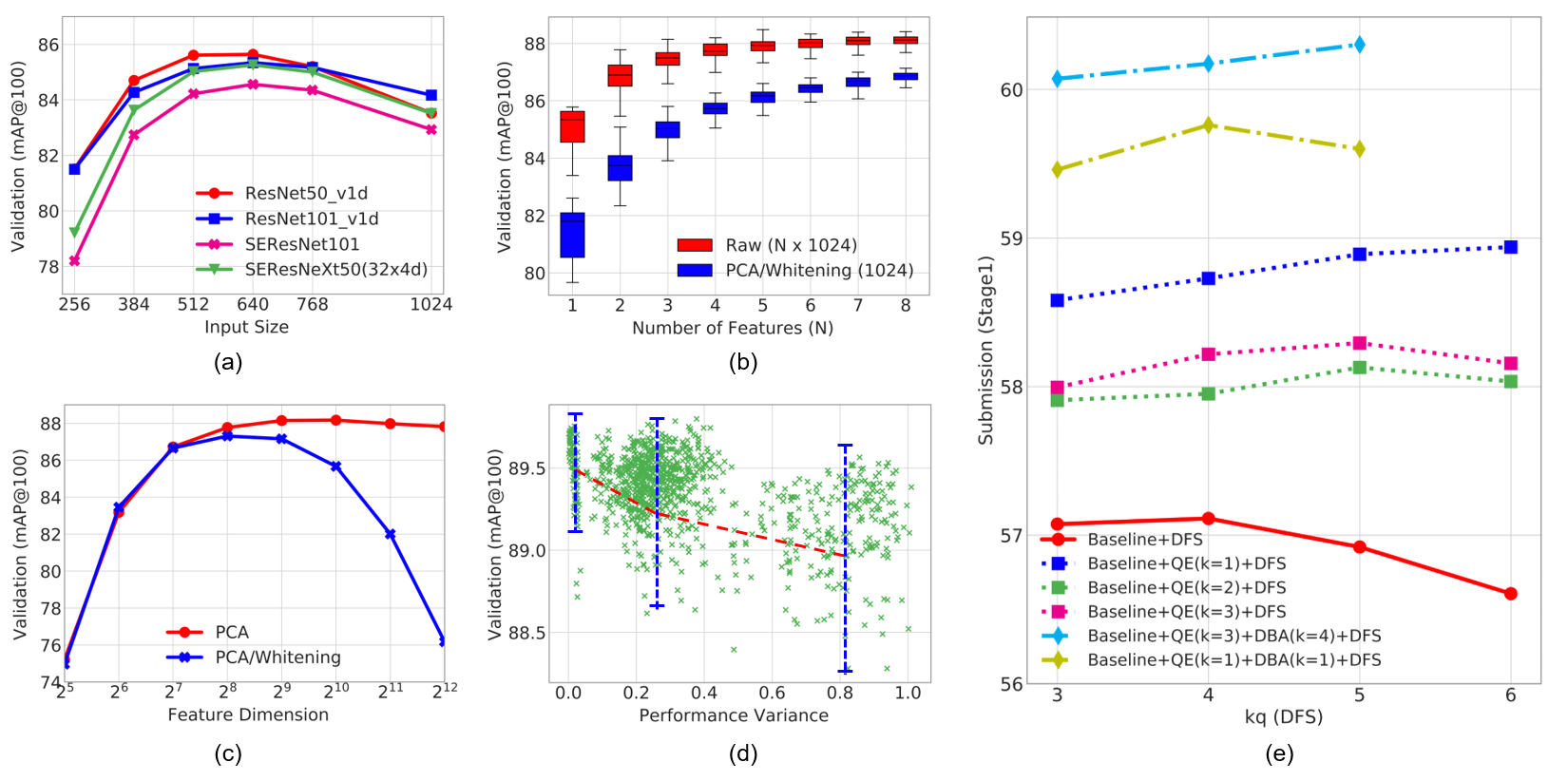}
\end{center}
   \caption{Experiments for performance tendency by varying each hyper-parameter. In (b) and (c), we did not apply DBA and QE, and performed PCA or PCA$_w$. In (d), each green point indicates a validation score of a feature ensemble, while the performance variance refers to the validation score variance among features of every single model. In (e), the performance is reported with the stage1 submission, while the baseline is a single ResNet50 model with Xent and Triplet loss.}
\label{fig:hyperparameter}
\end{figure*}

\vspace{-4mm}
\paragraph{Input size.}
We conducted experiments with an index-2 model in Table~\ref{Table:single_model_performance} by varying input size at the feature extraction step.
The result is shown in the Figure~\ref{fig:hyperparameter} (a).
It shows that performance rises as the input size increases as larger input sizes generate bigger feature maps in the convolutional neural network (CNN) models, which thus contain richer information.
However, the performance does not keep increasing indefinitely, starting to decrease at a certain point.

%%% Tricks
\section{Post-processing Tricks}
In this section, we investigate well-known post-processing tricks for instance-level image retrieval, including feature ensemble, database augmentation (DBA), query expansion (QE), and reranking.
Even though the effectiveness of each trick itself is well-proven in many studies, we found few papers that have empirically studied how these techniques should be mutually combined to maximize the performance.
We aim to figure out the mutual influence of combining tricks in different ways, and the sections below describe the details.

\subsection{Multiple Feature Ensemble}
The feature ensemble is a traditional method and the most representative technique for improving performance in many vision tasks~\cite{lin2018regional, kim2018attention, jun2019combination}.
Although feature ensembles can be seen as a simple concatenation of multiple features, there exists a point that is worth to consider: \textit{what features would be the best to be combined?}
Based on the single models we trained with various combinations of backbones, pooling methods, objectives, 1024-dimensional features were extracted, and a group of randomly chosen features was concatenated.
We examined how the performance varied according to the number of features concatenated.
In addition, we investigated whether the commonly used ``best only'' strategy, which picks the best performing features first, indeed guarantee better result in actual testing.

%Generally, ``the best only'' strategy which picks the best the single model performance is sorted

\subsection{DBA and QE}
Database-side augmentation replaces every feature point in the database with a weighted sum of the point's own value and those of its top $k$ nearest neighbors ($k$-NN)~\cite{turcot2009better, arandjelovic2012three}.
The purpose of DBA is to obtain more robust and distinctive image representations by averaging the feature points with the nearest neighbors.
We perform the weighted sum-aggregation of the descriptors with weights $W$ computed by,
\begin{equation}  \label{eq:weight}
W=logspace(a, b, n),
\end{equation}
where the $logspace$ function generates $n$ points between $10^a$ and $10^b$.

Similar to DBA, query expansion, introduced by Chum \textit{et al.}~\cite{chum2007total}, is a popular method of improving the quality of image retrieval by obtaining a richer representation of a query.
It retrieves top $k$ nearest neighbors from the database for each query and combines the retrieved neighbors with the original query.
This process is repeated with the number of necessity, and the final combined query is used to produce the ranked list of retrieved images.
More precisely, the weighted sum-aggregation of each query is performed with weights $W$ computed from Equation~\ref{eq:weight}.

%% Table 2: Feature Ensemble Performance
\begin{table*}[t!h!]
%\footnotesize
\centering
\begin{tabular}{cccccccccccc}
\hline
\multirow{2}{*}{Index} & \multirow{2}{*}{Combination} & \multicolumn{2}{c}{Concat.}      & \multicolumn{2}{c}{DBA}         & \multicolumn{2}{c}{DBA+QE}      & \multicolumn{2}{c}{DBA+PCA$_w$} & \multicolumn{2}{c}{DBA+QE+PCA$_w$} \\ \cline{3-12} 
                       &                              & Public         & Private        & Public         & Private        & Public         & Private        & Public         & Private        & Public           & Private         \\ \hline\hline
A                      & 0+3                          & 20.52          & 22.72          & 23.73          & 25.73          & 24.08          & 25.96          & 24.47          & 26.58          & 25.12            & 26.72           \\
B                      & 5+6                          & 18.76          & 21.56          & 23.46          & 25.22          & 23.83          & 25.47          & 23.61          & 25.78          & 24.05            & 26.07           \\ \hline
C                      & 0+2+4+8                      & 21.71          & 23.88          & 24.50          & 27.02          & 24.89          & 27.05          & 24.97          & 27.72          & 25.34            & 27.84           \\
D                      & 0+3+5+6                      & 21.05          & 23.43          & 24.50          & 26.77          & 25.10          & 26.85          & 24.96          & 27.34          & 25.49            & 27.47           \\
E                      & 0+5+6+7                      & 20.85          & 23.20          & 24.69          & 26.50          & 25.00          & 26.82          & 24.99          & 26.95          & 25.47            & 27.40           \\
F                      & 1+2+5+6                      & 20.56          & 22.79          & 24.02          & 26.15          & 24.27          & 26.38          & 24.39          & 26.71          & 24.76            & 26.83           \\
G                      & 4+6+7+8                      & 21.65          & 23.66          & 24.91          & 26.70          & 25.27          & 26.70          & 25.18          & 27.25          & 25.72            & 27.36           \\ \hline
H                      & 0+1+2+3+5+6                  & 21.04          & 23.30          & 24.33          & 26.44          & 24.62          & 26.70          & 24.79          & 27.05          & 25.07            & 27.23           \\
I                      & 0+2+3+4+7+8                  & 21.68          & 23.94          & 24.94          & \textbf{27.17} & \textbf{25.48} & \textbf{27.31} & 25.32          & \textbf{27.82} & \textbf{25.92}   & \textbf{27.96}  \\
J                      & 0+2+4+6+7+8                  & \textbf{22.03} & \textbf{23.98} & \textbf{25.03} & 27.14          & 25.42          & 27.26          & \textbf{25.44} & 27.70          & 25.77            & 27.83           \\ \hline
\end{tabular}
\caption{Performance (mAP@100) evaluation of the combination of multiple single models and each post-processing step. \textit{Concat.} row was performed by concatenating features without other post-processing. \textit{DBA} and \textit{DBA+QE} rows were evaluated by performing DBA and DBA+QE on every single feature and then concatenating features. \textit{DBA+PCA$_w$} and \textit{DBA+QE+PCA$_w$} rows followed the same process with \textit{DBA} and \textit{DBA+QE} rows with PCA$_w$ at the end. }
\label{Table:feature_ensemble_performance}
\end{table*}

%% Table 3: DFS + DELF
\begin{table}[t!h!]
\begin{adjustbox}{width=1.0\columnwidth,center}
% \footnotesize
\centering
\begin{tabular}{c c c c}
\hline
Method                     &   Public   &  Private   &    Diff.    \\
\hline \hline
Baseline                   &   25.88   &   27.94   &  -  \\
\hline
DFS (NN=1K, kq=4, ki=50)    &   25.94   &   27.60   &  \textcolor{red}{-0.34}  \\
DFS (NN=20K, kq=4, ki=50)   &   \textbf{26.67}   &   28.19   &  \textcolor{green}{+0.25}  \\
DFS (NN=40K, kq=4, ki=50)   &   26.61   &   \textbf{28.26}   &  \textcolor{green}{+0.32}  \\
DFS+SV (NN=1K, kq=4, ki=50) &   25.94   &   27.13   &  \textcolor{red}{-0.81}  \\
DFS+SV (NN=20K, kq=4, ki=50)&   26.64   &   27.71   &  \textcolor{red}{-0.23}  \\
DELF (kd=50)         &   25.88   &   27.92   &  \textcolor{red}{-0.02}  \\
DELF (kd=100)        &   25.85   &   27.87   &  \textcolor{red}{-0.07}  \\
DELF+D2R (kd=50)     &   25.95   &   27.89   &  \textcolor{red}{-0.05}  \\
DELF+D2R (kd=100)    &   25.96   &   27.88   &  \textcolor{red}{-0.06}  \\
\hline
\end{tabular}
\end{adjustbox}
\caption{Performance (mAP@100) evaluation of different methods of reranking. The \textit{Diff.} column indicate the difference between the private score and baseline. NN denotes the size of $k$-NN graph construction, when $kq$ and $ki$ are parameters for $k$-NN DFS search in query and index side, respectively. $kd$ in DELF is the number of candidates for reranking.}
\label{Table:dfs}
\end{table}

\subsection{PCA whitening}
In the retrieval task, whitening CNN-based descriptors have been promoted~\cite{sharif2014cnn, babenko2015aggregating} as they handle the problems arising from co-occurrence over-counting by jointly down-weighting co-occurrences~\cite{jegou2012negative}.
Typically, whitening is learned by a generative model in an unsupervised manner via principal component analysis (PCA) on an independent dataset.
We performed PCA whitening (PCA$_w$) with 4096-dimensional features from DBA and QE to produce 1024-dimensional features by using the implementation in the Scikit-learn API~\cite{pedregosa2011scikit}; then, we applied $l_2$-normalization again.

\subsection{Reranking}

Once the top $k$ candidates were retrieved, reranking the order of the retrieved candidates could improve performance.
The influence of reranking tricks is relatively minor compared with the previously mentioned tricks in that reranking is effective only if the ground-truth images were found in the top $k$ candidates.
Nevertheless, proper usage of reranking certainly improves performance.
We performed reranking using two distinctive methods: a graph search based on the global descriptor, and local matching based on the local descriptor.

\vspace{-4mm}
\paragraph{Graph search.} Diffusion (DFS)~\cite{yang2018efficient,iscen2017efficient} is a mechanism that captures the image manifold in the feature space.
It searches on the manifold efficiently based on a neighborhood graph of the dataset constructed offline.
This method improves the retrieval of small objects and cluttered scenes in particular, which fits the dataset domain of the  Google Landmark Challenge.
The performance increase using DFS is huge, and many state-of-the-art image retrieval papers use it as the last step in maximizing the score on the benchmark dataset.

\vspace{-4mm}
\paragraph{Local matching.} We use the spatial information of images for reranking.
Given two images, The correspondence match is extracted, and the number of inliers are counted using RANSAC~\cite{fischler1981random,chum2003locally}.
Because performing geometric verification on all possible pairs of query and index images is expensive for large-scale data, we applied local matching to only the top $kd$ candidates, which is the retrieved result for the global descriptor.
We used DELF~\cite{noh2017large} pre-trained on landmark dataset, and 1K local features were extracted from each image as described in the paper.
In the experiment, we found that unrelated images sometimes obtained more than 10 match score, which caused the performance drop.
To suppress this, we reranked the candidates only when the match score exceeded a certain threshold ($\sigma = 50$).

\subsection{Experimental Results}
Based on the trained single models, we have examined the effect of the aforementioned post-processing methods.
The performance of the feature combination and each post-processing step can be found in Table~\ref{Table:feature_ensemble_performance}.
Figure~\ref{fig:hyperparameter} shows experiments for values of the hyper-parameters of the respective tricks.

\vspace{-4mm}
\subsubsection{Multiple Feature Ensemble}
% \paragraph{Multiple Feature Ensemble.}
Figure~\ref{fig:hyperparameter} (a) shows how the performance varies according to the number of concatenated features.
We found a better result as the number of features was increased, but the gain becomes slighter with more features.
Considering the computational cost and the performance gain, we recommend using 4$\sim$6 features for the concatenation.
We also investigated whether the ``best only'' strategy was an optimal way of finding the best combination of features for the ensemble.
To test this, more than 1,400 combinations of concatenated features were constructed and evaluated on the validation set, as shown in Figure~\ref{fig:hyperparameter} (d).
% The performance variance means the mAP@100 variance of a group of models which are randomly chosen.
The result shows that there is a correlation between the low-performance variation of the models and better performance (red line).
However, we should not entirely trust the ``best only'' strategy because the variance of the points (blue line) is not negligible.

\vspace{-4mm}
\subsubsection{DBA/QE and PCA$_w$}
% \paragraph{DBA/QE and PCA$_w$.}
For processing features with DBA, QE, and PCA, we can think of two methods, depending on which trick is used first: (i) performing PCA on the concatenated feature, followed by DBA and QE; or (ii) performing the DBA and QE on each feature first, concatenate the features, and then apply PCA.
We chose the latter as it consistently achieved better results.
Note that the result in Table~\ref{Table:feature_ensemble_performance} are based on evaluation in this manner.

In the experiments, we found an interesting point: the tricks work differently depending on whether DBA and QE are applied to the feature earlier.
Table~\ref{Table:feature_ensemble_performance} shows that all combinations of concatenated features show a consistent performance increase when PCA and whitening are applied after DBA and QE.
This result is incompatible with the result of not applying DBA and QE in Figure~\ref{fig:hyperparameter} (b), which implies that dimensional reduction using PCA and whitening does worsen performance.
Similarly, we found conflicting results for the optimal feature dimensionality.
For PCA$_w$ when DBA and QE were used, 1024 was the optimal dimensionality for output, but the same parameter value degraded the performance significantly when using features without DBA and QE, as seen in Figure~\ref{fig:hyperparameter} (c).
Interestingly, DBA and QE enhance the quality of the feature representations, which also make them robust against dimensionality reduction.
As shown in Table~\ref{Table:feature_ensemble_performance}, the gain from DBA and QE is largest among all tricks.
Iterative DBA and QE perform augmentation $k$ times, but we used $k=1$ for both DBA and QE as it performed the best.
% Note that it may differ depending on the domain, or the type of trick applied beforehand.  

%%%%%%%%%%%%%%%%%%%%%%%%%%%%%%%%%%%%%%%%%%%%%%%%
%% Uncomment this Table for the final copy.
%%%%%%%%%%%%%%%%%%%%%%%%%%%%%%%%%%%%%%%%%%%%%%%%
\begin{table}[h]
% \footnotesize
\centering
\begin{tabular}{c c c c c}
\hline
Rank    &   Team   &   MeanPos  &  Public & Private \\
\hline \hline
1   &   smlyaka   &  31.8  &   35.69 & 37.23  \\
2   &   imagesearch   &  34.4  &   32.25 & 34.75 \\
3   &   Layer 6 AI   &  40.8  &   29.85 & 32.18 \\
4  &   bestfitting   &  37.8  &   28.26 & 31.41 \\
5  &   {[ods.ai]} n01z3   &  40.9  &   28.43 & 30.67 \\
6  &   learner   &  44.8  &   26.95 & 29.25  \\
7  &   CVSSP   &  42.0  &   26.44 & 27.97 \\
\hline \hline
\textbf{8}  &   \textbf{Ours}   &  \textbf{43.5}  &   \textbf{25.78} & \textbf{27.60}  \\
\hline \hline
9  &   VRGPrague   &  46.4  &   23.49 & 25.71 \\
10  &   JL   &  47.9  &   22.81 & 25.05  \\
\hline
\end{tabular}
\caption{Final results (mAP@100) for the top 10 teams on the Google Landmark Retrieval Challenge 2019. MeanPos is a mean position of a first relevant image. If there was no relevant image in the top 100, the position was listed as 101. The score was obtained from the challenge evaluation server.}
\label{Table:kaggle}
\end{table}

%%%%%%%%%%%%%%%%%%%%%%%%%%%%%%%%%%%%%%%%%%%%%%%%
%% USE THIS ONLY FOR REVIEW COPY.
%%%%%%%%%%%%%%%%%%%%%%%%%%%%%%%%%%%%%%%%%%%%%%%%
% \begin{table}[t]
% % \footnotesize
% \centering
% \begin{tabular}{c c c c c}
% \hline
% Rank    &   Team   &   MeanPos  &  Public & Private \\
% \hline \hline
% -   &   -   &  -  &   - & - \\
% -   &   -   &  -  &   - & - \\
% -   &   -   &  -  &   - & - \\
% -   &   -   &  -  &   - & - \\
% \hline \hline
% \textbf{hidden}  &   \textbf{due}   &  \textbf{to}  &   \textbf{blind} & \textbf{review}  \\
% \hline \hline
% -   &   -   &  -  &   - & - \\
% -   &   -   &  -  &   - & - \\
% -   &   -   &  -  &   - & - \\
% -   &   -   &  -  &   - & - \\
% -   &   -   &  -  &   - & - \\
% \hline
% \end{tabular}
% \caption{Final results (mAP@100) for the top 10 teams on the Google Landmark Retrieval Challenge 2019. MeanPos is a mean position of a first relevant image. If there was no relevant image in the top 100, the position was listed as 101. The score was obtained from the challenge evaluation server.}
% \label{Table:kaggle}
% \end{table}

\vspace{-4mm}
\subsubsection{Reranking}
% \paragraph{Reranking.}
Recently, the concept of detect-to-retrieve~\cite{teichmann2018detect} (D2R) has been proposed, and we used the landmark detector of \cite{teichmann2018detect} to detect and crop the region of interest.
The cropped region is used for local feature extraction with DELF~\cite{noh2017large}, which is listed as \textit{DELF Rerank+D2R} in Table~\ref{Table:dfs}.
Although DELF achieved a competitive result on the \textit{Oxford5k} and \textit{Paris6k} datasets, we observed very slightly increased or even worsened performance after reranking with DELF in large-scale datasets such as GLD.

We also explored DFS for reranking.
As shown in Figure~\ref{fig:hyperparameter} (e), DFS reranking with DBA/QE improved the performance by the hyper-parameter $kq$, while DFS reranking without DBA/QE was not helpful.
This shows that the features found by applying DBA/QE can construct better image manifolds in the feature space for graph searches.
Furthermore, we conducted experiments by combining DFS with spatial verification (SV), which is denoted as DFS+SV.
DFS+SV replaces a pairwise similarity measure of the cosine similarity between the global descriptors with the spatial matching score of the local descriptors obtained by DELF reranking.
Table~\ref{Table:dfs} shows that DFS improves the performance, while it worsens with SV.
As the number $k$ for $k$-NN graph construction was increased, the performance improved, but the additional computation slowed the process as a tradeoff.

\vspace{-4mm}
\subsubsection{Google Landmark Retrieval Challenge}
% \paragraph{Google landmark retrieval challenge.}
The Table~\ref{Table:kaggle} shows the final result on the Google Landmark Retrieval Challenge 2019.
For the final submission, we chose combination \textit{J} from Table~\ref{Table:feature_ensemble_performance} and applied DBA and QE with $k=1$ on each features.
The features were then concatenated and PCA$_w$ was applied with an output dimensionality of 1024.
Finally, the top 100 candidates were reranked using DFS+SV (NN=20K), which gave higher performance than using DFS only at this time.
Our pipeline ranked 8-th on the leaderboard by properly combining well-known retrieval tricks with simple inference models and proper use of hyper-parameters.
We did not use any complicated architecture or loss functions.
Note that our final submission score was improved after the challenge, as shown in Table~\ref{Table:dfs} and Table~\ref{Table:kaggle}.

%%% Conclusion
\section{Conclusion}
In this paper, we have examined the effectiveness of pre-processing and post-processing tricks on the large-scale dataset.
% The retrieval performance increases in each step of the pipeline based on the GLD.
The tricks such as dataset cleaning, feature ensembling, DBA/QE, PCA$_w$, and reranking by graph search and local feature matching were used for our pipeline.
We showed that both learning a good image representation and applying proper pre-processing and post-processing tricks are important, and those tricks can significantly boost the overall performance.
% We analysized the effectiveness of each steps in detail with extensive experiments, and finally obtained up to XX of mAP@100 increase.
Finally, we could obtain up to 10.24 of mAP@100 increase compared to a baseline model.

% \clearpage
{\small
\bibliographystyle{ieee}
\bibliography{egbib}
}

\end{document}